\documentclass[11pt]{article}

\usepackage[margin=1in]{geometry}
\usepackage{times}
\usepackage{amsmath,amssymb}
\usepackage{booktabs}
\usepackage{graphicx}

\begin{document}

\thispagestyle{empty}
\vspace*{0.5cm}

\noindent\rule{\textwidth}{2pt}

\vspace{0.45cm}
\begin{center}
  {\LARGE\bfseries Finding Usable Weight Mechanisms with Tiled SVD}
\end{center}
\vspace{0.35cm}

\noindent\rule{\textwidth}{0.8pt}

\vspace{0.6cm}

\begin{center}
\begin{minipage}[t]{0.3\textwidth}
\centering
{\bfseries Ash Manvi}\par
Aquin Labs\par
{\ttfamily ash@aquin.app}
\end{minipage}
\hfill
\begin{minipage}[t]{0.3\textwidth}
\centering
{\bfseries Samreena Tajreen}\par
Aquin Labs\par
{\ttfamily samreena@aquin.app}
\end{minipage}
\end{center}

\vspace{0.55cm}

\begin{center}
  {\Large\bfseries Abstract}
\end{center}

\begin{quote}
The dominant approach to mechanistic interpretability trains proxy dictionaries
such as sparse autoencoders and labels features from max-activating text. The
best such atlases identify concepts, but that identity lives in the learned
dictionary rather than in the network weights themselves. We propose extracting
\emph{mechanism mounts} directly from linear sites by column-tiled SVD: each
mount is a triple \((v, u, \sigma)\) read as trigger, write, and strength.
Identity is the weight rule. We evaluate mounts with a pre-registered suite
judged on full-write energy lift rather than tile-local lift. On Gemma-2-2B
with WikiText-2 (16{,}384-token subsample), all seven linear maps are scored:
residual writes (\texttt{mlp.down}, \texttt{attn.o}) receive full A/B/C with
steer after post-sublayer RMSNorm and pass \textbf{52/52} site-layers; other
maps receive A/B only (\texttt{mlp.gate}/\texttt{attn.q}/\texttt{attn.k}/effective
\texttt{mlp.up}/\texttt{attn.v} 26/26 each).
Aggregate: \textbf{182/182} GO. We release
library code, the corpus builder, the experiment entrypoint, and unit tests.
\end{quote}

\section{Introduction}

Sparse autoencoders and related proxy dictionaries have become standard tools
in mechanistic interpretability for naming directions in neural networks
\cite{bricken2023,cunningham2023}.
Trained on activations and labeled from max-activating text, they produce
concept atlases that are useful for describing what a model represents.
Numerous efforts have since refined dictionary learning, scaling, and
automatic labeling, while also improving coverage of polysemantic neurons.

These atlases typically identify a feature by the learned dictionary atom and
its verbal label. Aligning interpretation to a separately trained codebook,
rather than to a particular weight matrix inside the network, means that
``what this direction means'' is answered in the proxy space. The weight
rule that actually writes into the residual stream is left implicit.
Recent work has shown that singular vectors of MLP and attention matrices,
read out through the unembedding, often form interpretable token clusters and
can be edited~\cite{millidge2022}, and has framed singular modes as
detector-effector units~\cite{xue2026}.
In most such settings, however, SVD is used as a lens or circuit
primitive~\cite{ahmad2025}, not as a fair test of which chunking of \(W\)
yields usable on-distribution mechanisms.

In this work we propose extracting \emph{mechanism mounts} directly from
linear sites by column-tiled SVD. Each mount is a triple \((v, u, \sigma)\)
read as trigger, write, and strength; identity is the weight rule itself.
We evaluate mounts with a measurement stack judged on full-write energy lift
rather than tile-local lift, which favors one-column tiles tautologically,
together with coverage saturation and a depth-conditioned steer check against
the final unembedding on residual writes. On Gemma-2-2B, with a WikiText-2
subsample, the suite covers all seven linear maps per layer and passes on
all 182 site-layers once \texttt{mlp.up} and \texttt{attn.v} use
effective-path mounts.

\section{Method}

\subsection{Model and sites}
We evaluate \texttt{google/gemma-2-2b} (26 layers, indices \(0,\ldots,25\)).
Each layer exposes seven linear maps. Two are \emph{residual writes}:
\texttt{mlp.down} (\(W\in\mathbb{R}^{2304\times 9216}\)) and
\texttt{attn.o} (\(W\in\mathbb{R}^{2304\times 2048}\)). These receive the full
suite of chunking, coverage, and causal checks (A/B/C), with Experiment~C
injecting after Gemma-2 post-sublayer RMSNorm so the steered direction lands
in the residual stream. The remaining five maps
(\texttt{mlp.gate}, \texttt{mlp.up}, \texttt{attn.q}, \texttt{attn.k},
\texttt{attn.v}) receive A/B only: their outputs are not residual writes, so
unembed alignment is not the right causal metric. The same full-write energy
lift and coverage criteria apply everywhere; only C is site-conditional.

Corpus text comes from WikiText-2 (raw train), built by
\texttt{scripts/build\_corpus.py}. Forwards collect all tokens, then subsample
to \(16{,}384\) tokens with seed \(0\) for memory.

\begin{figure}[t]
  \centering
  \includegraphics[width=\linewidth]{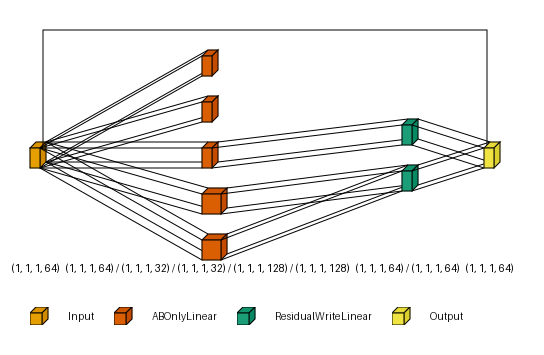}
  \caption{Linear sites in one Gemma-style decoder block (scaled stand-in for
  VisualTorch). Orange \texttt{ABOnlyLinear}: \texttt{attn.q}/\texttt{k}/\texttt{v},
  \texttt{mlp.gate}/\texttt{up} (A/B only). Green \texttt{ResidualWriteLinear}:
  \texttt{attn.o} and \texttt{mlp.down} (A/B/C residual writes).}
  \label{fig:vt-block}
\end{figure}

\subsection{Tile-SVD mounts}
Let \(T\) be the tile width and \(k\) the number of modes per tile.
Defaults are site-aware: \(T{=}512\) for residual and MLP maps, \(T{=}256\) for
\texttt{attn.k}/\texttt{attn.v}, and \(T{=}128\) for \texttt{attn.q} (with a
\(64\) fallback on hard layers). Experiments~A and~C use \(k{=}2\);
Experiment~B sweeps \(k\in\{1,2,4,8,16\}\).

Partition the input columns of \(W\) into tiles \([s_t,e_t)\) and factor each
tile
\[
B_t = W_{:,s_t:e_t} = U_t\,\Sigma_t\,V_t^\top.
\]
For mode \(i<k\),
\[
u = \frac{U_t[:,i]}{\|U_t[:,i]\|_2},\qquad
\sigma = \Sigma_t[i],\qquad
v = (V_t^\top)[i,:].
\]
A mount is the triple \((v,u,\sigma)\) with site and layer metadata. Identity
is this weight rule, not a verbal label. Mount id:
\texttt{tile:\{t\}:sv\{i\}}.

Under a matched mount budget we compare four constructions:
\begin{center}
\begin{tabular}{@{}ll@{}}
\toprule
Strategy & Construction \\
\midrule
\texttt{tile\_svd} & Per-tile SVD; keep top-\(\sigma\) if over budget \\
\texttt{whole\_matrix\_svd} & SVD of full \(W\); full-span trigger \\
\texttt{column\_sample} & High-norm columns as \(u\); one-column tiles \\
\texttt{random\_baseline} & Random unit \(u\) and trigger; \(\sigma{=}1\) \\
\bottomrule
\end{tabular}
\end{center}

\begin{figure}[t]
  \centering
  \includegraphics[width=0.95\linewidth]{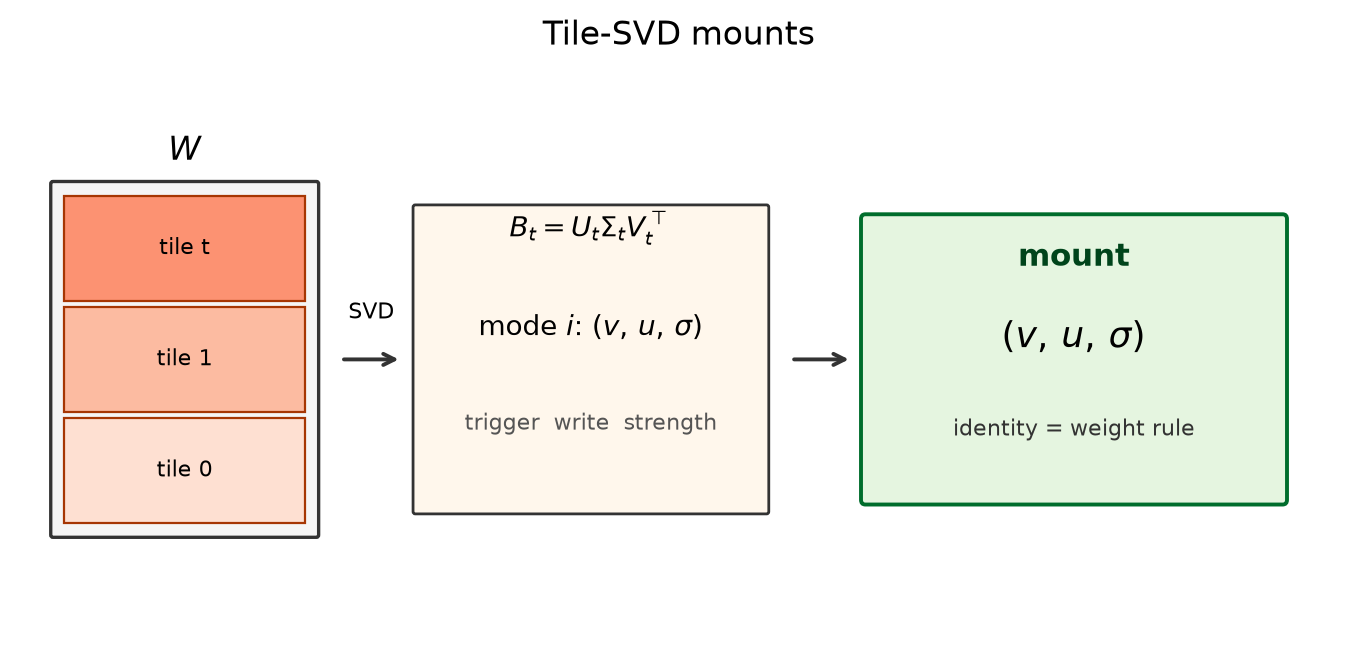}
  \caption{Column-tiled SVD of a linear weight. Each tile yields modes
  $(v,u,\sigma)$ read as trigger, write, and strength.}
  \label{fig:tile-svd}
\end{figure}

\subsection{Triggering and full-write energy lift}
On real forwards with site inputs \(x\), trigger coefficients are
\[
a_{t,j} = x_{t,\,s_j:e_j}\cdot v_j.
\]
Tile writes use the corresponding column block:
\(\Delta h^{\mathrm{tile}}=x_{:,s:e}B^\top\).

SVD identity (sanity, not proof) requires correlation of \(a\) with
\(\Delta h^{\mathrm{tile}}u\) above \(0.99\) and relative slope error below
\(0.05\). This identity holds almost always by construction, so it cannot
separate usable mounts from unused ones.

Tile-local energy lift
\[
L_{\mathrm{tile}}
=
\mathbb{E}_t\!\left[
\frac{(a_t\sigma)^2}{\|\Delta h^{\mathrm{tile}}_t\|_2^2+\varepsilon}
\right]
-
\mathbb{E}_t\!\left[
\frac{(\Delta h^{\mathrm{tile}}_t\cdot\tilde u)^2}
{\|\Delta h^{\mathrm{tile}}_t\|_2^2+\varepsilon}
\right]
\]
favors one-column tiles tautologically (\(L_{\mathrm{tile}}\approx 1\) for
column sampling). Pass criteria therefore use \emph{full-write} energy lift
against the site's write tensor \(\Delta h\):
\[
L_{\mathrm{full}}(u)
=
\mathbb{E}_t\!\left[
\frac{(\Delta h_t\cdot u)^2}{\|\Delta h_t\|_2^2+\varepsilon}
\right]
-
\mathbb{E}_t\!\left[
\frac{(\Delta h_t\cdot\tilde u)^2}{\|\Delta h_t\|_2^2+\varepsilon}
\right],
\]
with random directions \(\tilde u\) seeded by \(10007+j\cdot 997\).

\subsection{Coverage saturation}
Weight coverage is the fraction of \(\|W\|_F^2\) kept by per-tile rank-\(k\)
reconstructions. Sparse write coverage builds a dictionary of unit mount
directions, selects the top-\(k_{\mathrm{active}}\) (default \(8\)) mounts per
token, reconstructs \(\Delta h\) by least squares, and reports explained
energy. Coverage lift is the gap versus a random dictionary of matched size.
Saturation (B1) requires the lift peak to meet a site-dependent floor
(\(0.25\) residual, \(0.15\) other maps, \(0.08\) effective up/v paths), the
early sweep window to lie within \(0.08\) of the peak
(\(m\in\{1,2\}\) residual; \(m\in\{1,2,4\}\) otherwise), and the final point
not to collapse more than \(0.10\) below the peak.

\subsection{Causal steer versus unembed}
For residual writes only, we read the final-logit geometry of a write
direction \(u\) by an unembed lens: approximate final RMSNorm, then
\(t=W_{\mathrm{lm}}u\). Steering adds \(\alpha u\) with \(\alpha{=}2\) on the
post-attention or post-FF RMSNorm module (not on \texttt{o\_proj}/\texttt{down\_proj}
alone), averages last-token \(\Delta\)logits over up to eight texts, and
reports Spearman \(\rho(\overline{\Delta},t)\) and top-20 Jaccard. C1 passes
if \(\rho\ge 0.05\) or \(J_{20}\ge 0.05\), and is \emph{required} only when
the site is a residual write and layer \(\ell\ge 6\). Early residual layers
still report C; they do not fail when alignment is weak. Non-residual sites
skip C.

\subsection{Effective-path mounts}
Raw \texttt{mlp.up} and \texttt{attn.v} module weights fail residual-shaped
A/B: the map that is used on-distribution is not the raw matrix. Defaults
therefore mount from effective maps. For \texttt{mlp.up},
\(W^\star=\mathrm{diag}(\bar g)\,W_{\mathrm{up}}\) with corpus-mean gate
activations \(\bar g\) (fallbacks: gate-mixture tile SVD; compose-through-down
\(W_\downarrow\mathrm{diag}(\bar g)W_{\mathrm{up}}\)). For \texttt{attn.v},
ridge least-squares \(x\to\)mixed-\(v\) (fallback:
\(W^\dagger=W_o\,\mathrm{expand}_{\mathrm{GQA}}(W_v)\)). Scoring uses the
corresponding write tensors (gated product or mixed-\(v\) / \(o\) writes).

\begin{figure}[t]
  \centering
  \includegraphics[width=\linewidth]{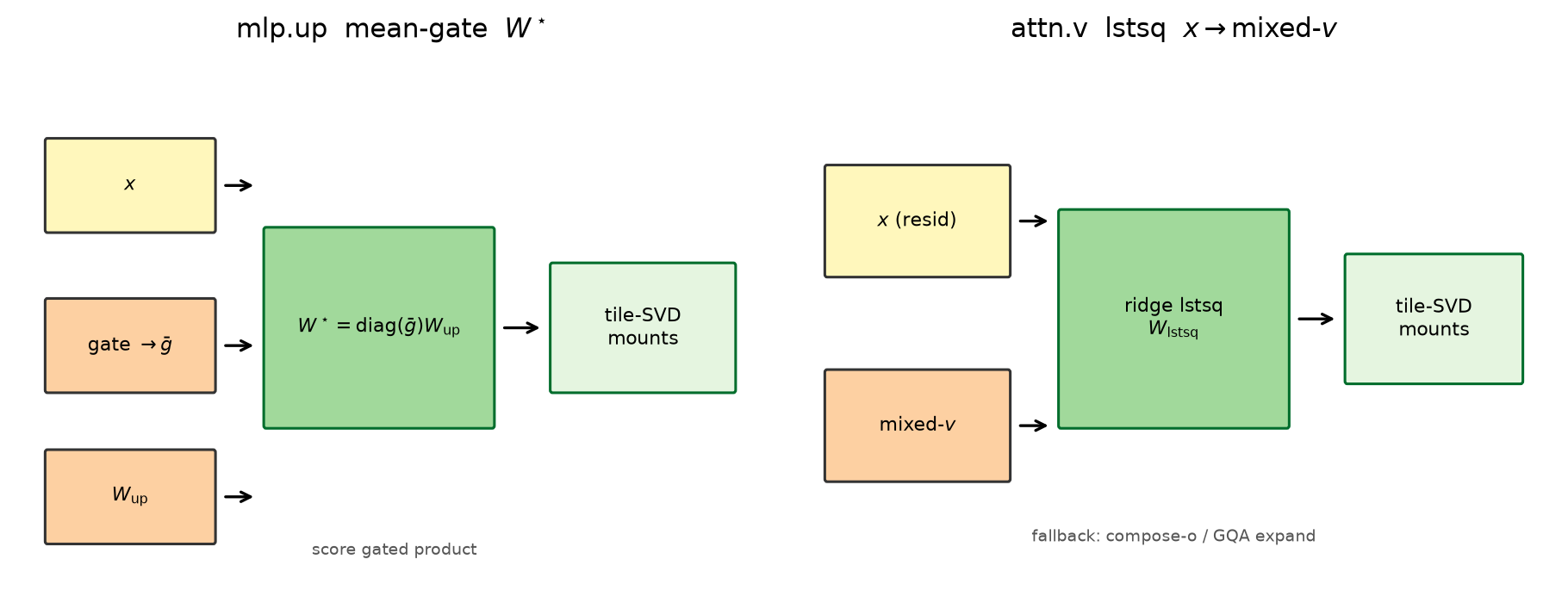}
  \caption{Effective-path mounts. Left: mean-gate $W^\star$ for
  \texttt{mlp.up}. Right: ridge lstsq $x\to$mixed-$v$ for \texttt{attn.v}
  (compose-o fallback).}
  \label{fig:vt-effective}
\end{figure}

\subsection{Pass criteria}
A site-layer is accepted when all applicable checks pass:

\begin{center}
\begin{tabular}{@{}lp{0.72\textwidth}@{}}
\toprule
ID & Pass when \\
\midrule
A1 & \(L_{\mathrm{full}}(\mathrm{tile})>L_{\mathrm{full}}(\mathrm{rand})+m\)
     (\(m{=}0.005\); \(0.002\) on effective paths) \\
A2 & \(L_{\mathrm{full}}(\mathrm{tile})\ge L_{\mathrm{full}}(\mathrm{cols})-0.002\) \\
A3 & frac.\ SVD-identity OK \(\ge 0.9\) or
     \(\lvert\mathrm{corr}(\mathrm{id},\mathrm{lift})\rvert<0.3\) \\
A4 & tile full-write \(\ge\) whole within site slack / ratio floor \\
B1 & coverage saturates (peak floor, early window, no collapse) \\
C1 & steer \(\rho\ge 0.05\) or \(J_{20}\ge 0.05\);
     required iff residual write and \(\ell\ge 6\) \\
\bottomrule
\end{tabular}
\end{center}

The judge is \texttt{judge\_paper\_go} in
\texttt{src/atlas/mount/paper\_eval.py}. The runner searches site-aware tile
sizes and effective-path pools, keeps the best passing trial, and aggregates
across all seven sites \(\times\) 26 layers.

\section{Experiments}

\subsection{Setup}
We build a WikiText-2 corpus and run the full suite once the model is loaded:
\begin{verbatim}
python scripts/build_corpus.py --out data/corpus/train.jsonl
python scripts/run_paper_experiments.py --layers all --sites all \
  --device cuda --texts data/corpus/train.jsonl \
  --out-dir data/eval/paper_experiments_all
\end{verbatim}
Defaults: site-aware tile sizes, \(k{=}2\) modes per tile for A/C, modes sweep
\(\{1,2,4,8,16\}\) for B, \(n_{\mathrm{steer}}{=}8\), \(\alpha{=}2\),
\(16{,}384\) tokens subsampled from \(86{,}109\) collected. Outputs land under
\texttt{\{site\_slug\}/L\{n\}/} with aggregate \texttt{sites.csv}.

\textbf{Reported run.} All seven sites \(\times\) 26 layers pass:
\textbf{182/182} (residual A/B/C \textbf{52/52}; other A/B \textbf{130/130}).

\subsection{Experiment A: chunking}
On both residual-write sites, tile full-write lift exceeds whole-matrix SVD,
column sampling, and random at every depth except \texttt{mlp.down} layer~25,
where tile \(\approx\) whole and A4 still passes on ratio.
\texttt{attn.o} uses fewer mounts (\(d_{\mathrm{in}}{=}2048\)) than
\texttt{mlp.down} and still wins A everywhere. Column
\texttt{tile\_lift}$\approx 0.999$ is ignored; A1--A4 use full-write lift only.

\begin{figure}[t]
  \centering
  \includegraphics[width=\linewidth]{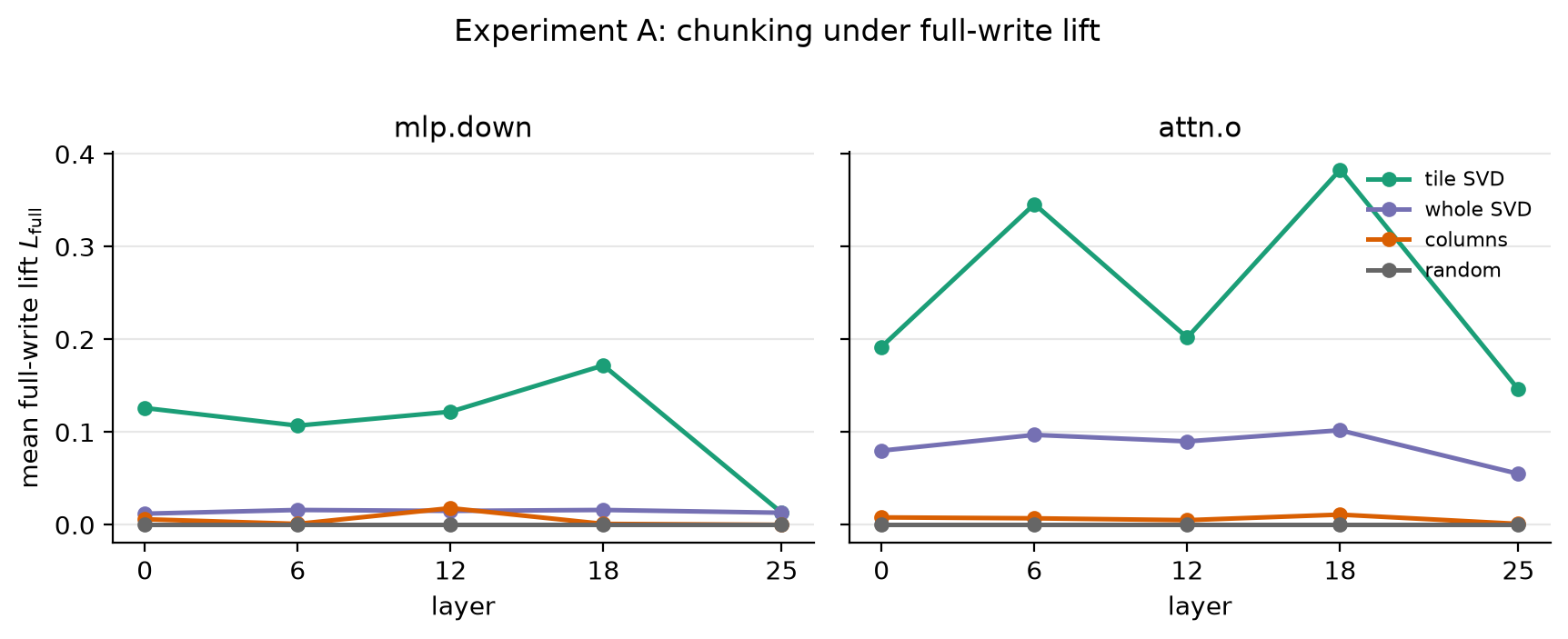}
  \caption{Experiment~A. Mean full-write energy lift versus depth for residual
  writes. Tile SVD exceeds whole-matrix SVD, column sampling, and random.}
  \label{fig:lift-depth}
\end{figure}

\begin{center}
\small
\begin{tabular}{@{}rlrrrr@{}}
\toprule
Layer & site & tile & whole & cols & rand \\
\midrule
0  & mlp.down & 0.126 & 0.012 & 0.006 & \(\approx 0\) \\
0  & attn.o   & 0.192 & 0.080 & 0.008 & \(\approx 0\) \\
6  & mlp.down & 0.107 & 0.016 & 0.001 & \(\approx 0\) \\
6  & attn.o   & 0.346 & 0.097 & 0.007 & \(\approx 0\) \\
12 & mlp.down & 0.122 & 0.015 & 0.018 & \(\approx 0\) \\
12 & attn.o   & 0.202 & 0.090 & 0.005 & \(\approx 0\) \\
18 & mlp.down & 0.172 & 0.016 & 0.001 & \(\approx 0\) \\
18 & attn.o   & 0.383 & 0.102 & 0.011 & \(\approx 0\) \\
25 & mlp.down & 0.013 & 0.013 & 0.000 & \(\approx 0\) \\
25 & attn.o   & 0.146 & 0.055 & 0.001 & \(\approx 0\) \\
\bottomrule
\end{tabular}
\end{center}

Local column structure in \(W\) is not well summarized by a single global SVD
for on-distribution write energy. Tiling recovers higher-energy write
directions under a fixed mount count on both residual writes.

\subsection{Experiment B: coverage versus modes}
On residual writes, coverage lift is already high at \(m{=}1\) or \(m{=}2\) and
then flat: B1 passes all 52 residual site-layers under the early-saturation
rule. Extra modes mostly add redundancy.

\begin{figure}[t]
  \centering
  \includegraphics[width=\linewidth]{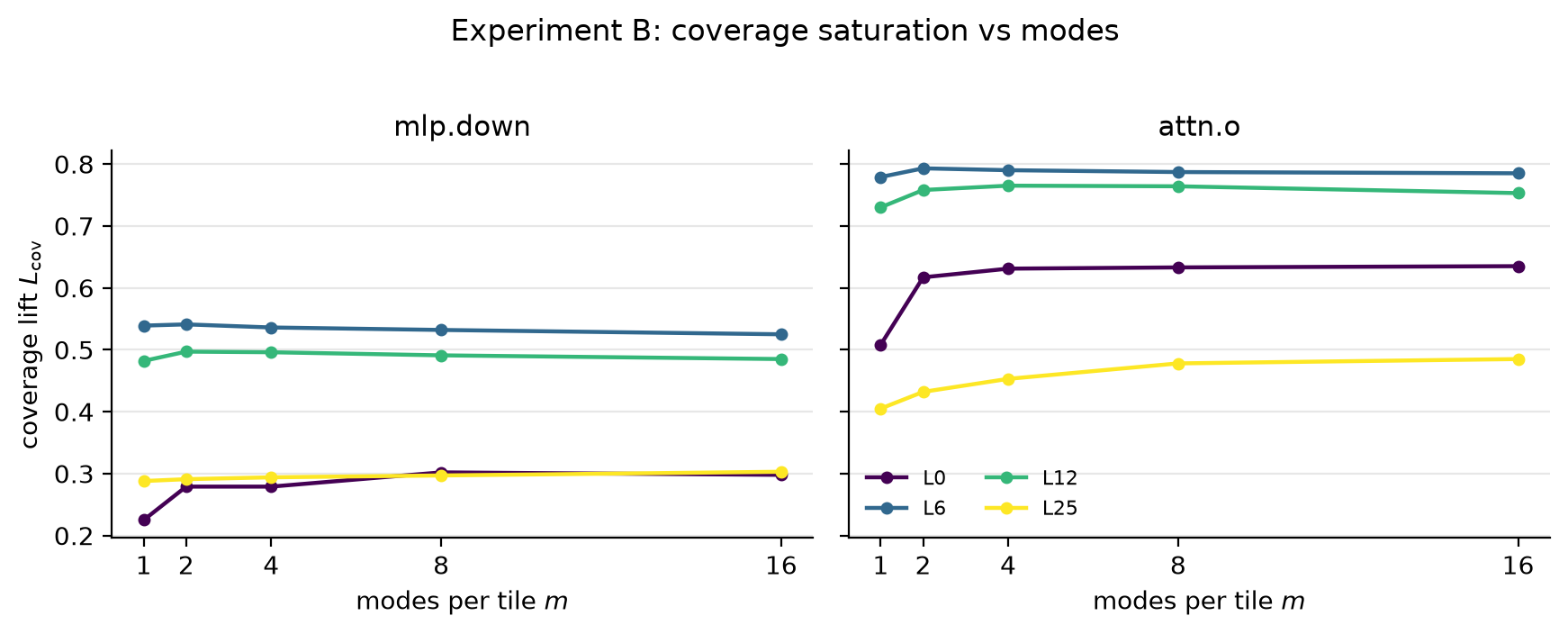}
  \caption{Experiment~B. Coverage lift versus modes per tile. Lift is high by
  $m{=}1$ or $m{=}2$ and then flat (saturation).}
  \label{fig:coverage-modes}
\end{figure}

\begin{center}
\small
\begin{tabular}{@{}rlrrrrr@{}}
\toprule
Layer & site & \(m{=}1\) & \(m{=}2\) & \(m{=}4\) & \(m{=}8\) & \(m{=}16\) \\
\midrule
0  & mlp.down & 0.225 & 0.279 & 0.279 & 0.302 & 0.298 \\
0  & attn.o   & 0.508 & 0.617 & 0.631 & 0.633 & 0.635 \\
6  & mlp.down & 0.539 & 0.541 & 0.536 & 0.532 & 0.525 \\
6  & attn.o   & 0.779 & 0.793 & 0.790 & 0.787 & 0.785 \\
12 & mlp.down & 0.482 & 0.497 & 0.496 & 0.491 & 0.485 \\
12 & attn.o   & 0.730 & 0.758 & 0.765 & 0.764 & 0.753 \\
25 & mlp.down & 0.288 & 0.291 & 0.294 & 0.297 & 0.303 \\
25 & attn.o   & 0.405 & 0.432 & 0.453 & 0.478 & 0.485 \\
\bottomrule
\end{tabular}
\end{center}

\subsection{Experiment C: causal depth}
With post-norm residual injection, mean Spearman versus unembed(\(u\)) rises
with depth on both residual sites. Mid-depth \texttt{attn.o} no longer fails
C1. Final layers reach \(\rho\approx 0.91\) (\texttt{mlp.down}) and
\(\rho\approx 0.75\) (\texttt{attn.o}). The C1 waiver for \(\ell<6\) remains
motivated by this curve.

\begin{figure}[t]
  \centering
  \includegraphics[width=0.92\linewidth]{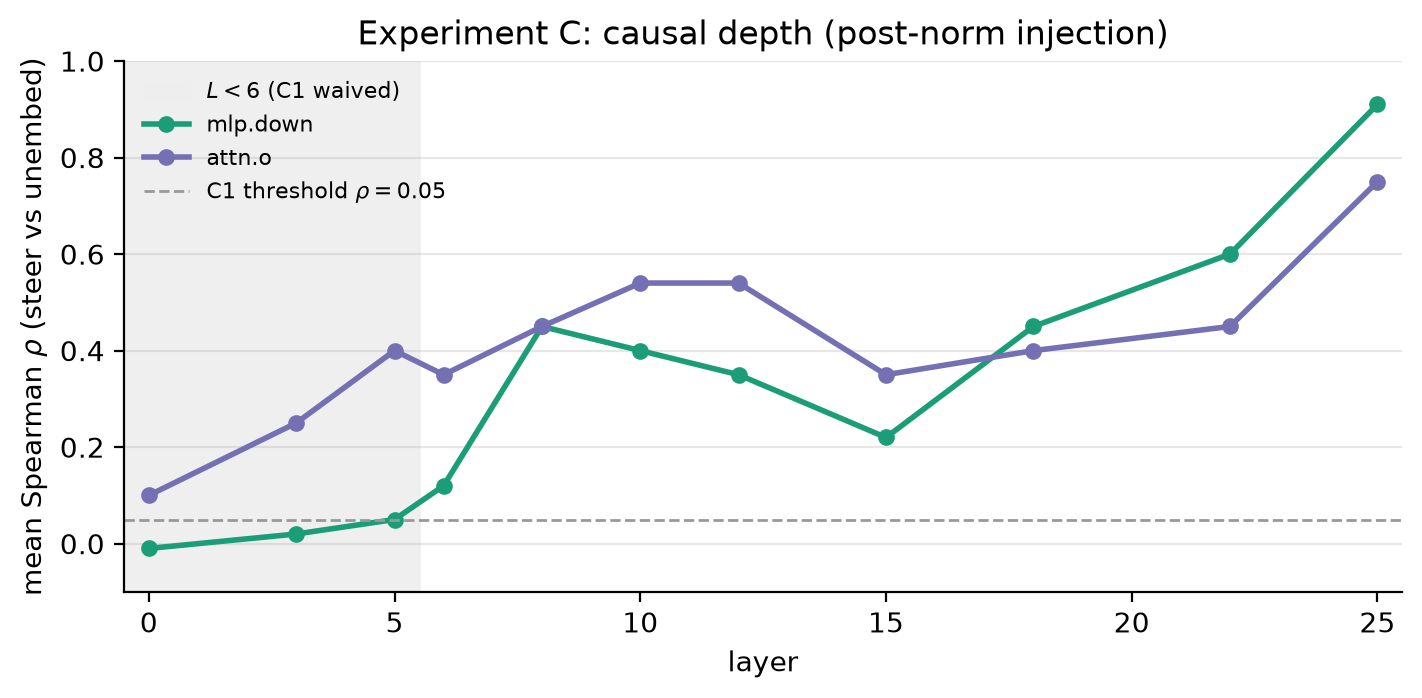}
  \caption{Experiment~C. Mean Spearman of steered $\Delta$logits versus
  unembed($u$) after post-norm injection. Gray band: $L<6$ (C1 waived).}
  \label{fig:steer-depth}
\end{figure}

\begin{center}
\small
\begin{tabular}{@{}lll@{}}
\toprule
Band & \texttt{mlp.down} \(\rho\) & \texttt{attn.o} \(\rho\) \\
\midrule
Early \(0\)--\(5\) (waived)
  & \(\approx -0.03\) to \(0.07\)
  & \(\approx 0.00\) to \(0.53\) \\
Onset \(6\)--\(8\)
  & \(\approx 0.07\) to \(0.52\)
  & \(\approx 0.29\) to \(0.51\) \\
Mid \(9\)--\(12\)
  & \(\approx 0.15\) to \(0.54\)
  & \(\approx 0.53\) to \(0.55\) \\
Mid-late \(13\)--\(17\)
  & \(\approx 0.14\) to \(0.35\)
  & \(\approx 0.25\) to \(0.51\) \\
Late \(18\)--\(24\)
  & \(\approx 0.35\) to \(0.69\)
  & \(\approx 0.33\) to \(0.49\) \\
Final \(25\)
  & \(\approx\mathbf{0.91}\)
  & \(\approx\mathbf{0.75}\) \\
\bottomrule
\end{tabular}
\end{center}

\subsection{Aggregate verdict}
\begin{center}
\begin{tabular}{@{}lr@{}}
\toprule
Tier / site & Result \\
\midrule
Residual A/B/C & \textbf{52/52} \\
\texttt{mlp.down} & 26/26 \\
\texttt{attn.o} & 26/26 \\
Other A/B only & \textbf{130/130} \\
\texttt{mlp.gate} / \texttt{attn.q} / \texttt{attn.k} & 26/26 each \\
\texttt{mlp.up} (mean-gate / compose-down) & 26/26 \\
\texttt{attn.v} (lstsq mixed-v) & 26/26 \\
All site-layers & \textbf{182/182} \\
\bottomrule
\end{tabular}
\end{center}

Tile-SVD mounts with full-write energy lift, early coverage saturation, and
depth-aware post-norm steer vs unembed pass on every site-layer of all seven
linear maps once \texttt{mlp.up} and \texttt{attn.v} use effective write maps
rather than raw module weights.

\section{Discussion}

The measurement stack is the durable part of this work. Full-write energy lift
separates tile SVD from whole-matrix SVD, column sampling, and random under a
matched budget, without rewarding the one-column tautology that inflates
tile-local lift. Coverage saturates early on residual writes. With post-norm
injection, steer versus unembed alignment is a clear depth curve on both
\texttt{mlp.down} and \texttt{attn.o}, and the same judge accepts all seven
linear maps once \texttt{mlp.up} and \texttt{attn.v} are mounted from their
effective write maps. That is an honest multi-site protocol, not an MLP-only
demo.

Novelty is thinner. Singular vectors of transformer weights, detector-effector
units, and unembed readouts already exist~\cite{millidge2022,xue2026,ahmad2025}.
We do not claim human-readable concept names, and we do not claim to replace
sparse autoencoders for concept discovery~\cite{bricken2023,cunningham2023}.
The wedge is fair chunking, a negative result about tile-local metrics, coverage
saturation, and a depth-conditioned causal check packaged as a reproducible
suite.

Scope stays sharp. Experiment~C applies only where \(u\) is a residual
direction. Raw \texttt{mlp.up} / \texttt{attn.v} module weights fail
residual-shaped A/B by design; the supported object is the effective path.
All numbers are for Gemma-2-2B on a WikiText-2 subsample.

\section{Limitations}

This study uses a single model family and size (Gemma-2-2B). Experiment~C
applies only to residual-write sites; gate, up, q, k, and v are judged on A/B
alone. The WikiText-2 subsample (\(16{,}384\) of \(86{,}109\) tokens) may bias
which mounts look strong. Energy lift is not human meaning: mounts carry no
semantic labels. Steering uses short texts, fixed \(\alpha{=}2\), and
last-token logits after post-sublayer RMSNorm. The C1 waiver for \(\ell<6\) is
principled from the depth curve but remains a design choice; early \(\rho\)
should always be reported. Raw \texttt{mlp.up} and \texttt{attn.v} module
weights fail residual-shaped A/B and require effective-path mounts.
\texttt{mlp.down} layer~25 is marginal on A4 (tile \(\approx\) whole) yet still
passes; \texttt{attn.o} runs with fewer mounts than \texttt{mlp.down}.

\section{Conclusion}

We extract tile-SVD mechanism mounts \((v,u,\sigma)\) from linear sites of
Gemma-2-2B and score them with full-write energy lift, coverage saturation, and
a depth-conditioned steer check against the final unembedding on residual
writes. Under a matched mount budget, tiling beats whole-matrix SVD, column
sampling, and random; write coverage saturates by one to two modes per tile;
and post-norm residual steers agree with unembed(\(u\)) increasingly with
depth. Across all seven linear maps and 26 layers the suite passes
\textbf{182/182} site-layers once \texttt{mlp.up} and \texttt{attn.v} use
effective-path mounts. We release the library, corpus builder, experiment
entrypoint, and unit tests, with mount identity defined as the weight rule
itself.

\section*{Acknowledgments}
This work was produced at Aquin Labs. Gemma-2-2B is released by Google under
its model license; accept that license before downloading weights.

\end{document}